\documentclass[sigconf]{acmart}
\AtBeginDocument{%
  }

\setcopyright{acmlicensed}
\copyrightyear{2025}
\acmYear{2025}
\acmDOI{XXXXXXX.XXXXXXX}
\acmConference[CIKM '25]{Conference on Information and Knowledge Management}{November 10--14, 2025}{Seoul, Korea}
\acmISBN{}

\usepackage[para]{footmisc}

\begin{document}

\title{Safeguarding Generative AI Applications in Preclinical Imaging through Hybrid Anomaly Detection }

\author{Jakub Binda}
\affiliation{%
  \institution{Institute of Informatics, University of Warsaw}
  \institution{Alethia XAI Sp. z o.o.}
  \city{Warsaw}
  \country{Poland}
}

\author{Valentina Paneta}
\author{Vasileios Eleftheriadis}
\affiliation{%
  \institution{BIOEMTECH}
  \city{Athens}
  \country{Greece}}

\author{Hongkyou Chung}
\affiliation{%
  \institution{School of Law, Seoul National University}
  \institution{Shin \& Kim LLC,}
  \city{Seoul}
  \country{Republic of Korea}}

\author{Panagiotis Papadimitroulas}
\affiliation{%
  \institution{BIOEMTECH, Athens}
  \institution{Medical Informatics Laboratory, School of Medicine, University of Thessaly, Larissa}
  \country{Greece}}

\author{Neo Christopher Chung}
\affiliation{%
  \institution{Institute of Informatics, University of Warsaw}
  \institution{Alethia XAI Sp. z o.o.}
  \city{Warsaw}
  \country{Poland}
}

\renewcommand{\shortauthors}{Binda et al.}

\begin{abstract}

Generative AI holds great potentials to automate and enhance data synthesis in nuclear medicine. However, the high-stakes nature of biomedical imaging necessitates robust mechanisms to detect and manage unexpected or erroneous model behavior. We introduce development and implementation of a hybrid anomaly detection framework to safeguard GenAI models in BIOEMTECH's \emph{eyes}\textsuperscript{\texttrademark}  systems. Two applications are demonstrated: Pose2Xray, which generates synthetic X-rays from photographic mouse images, and DosimetrEYE, which estimates 3D radiation dose maps from 2D SPECT/CT scans. In both cases, our outlier detection (OD) enhances reliability, reduces manual oversight, and supports real-time quality control. This approach strengthens the industrial viability of GenAI in preclinical settings by increasing robustness, scalability, and regulatory compliance.
  
\end{abstract}




\maketitle

\section{Introduction}


Generative adversarial networks (GANs) are a powerful class of deep learning models that train a generator to capture the data distribution via an adversarial game \cite{goodfellow2020gan, vox2vox}. These advances are gradually being adopted in nuclear medicine \cite{vasileios2025, fysikopoulos2021}. Since such synthetic images are used to aid downstream data inspection and analysis, generative AI (GenAI) systems in preclinical imaging must be robust to unexpected inputs.

Anomaly detection aims to identify inputs that deviate from the training distribution \cite{kennedy2025-outlier, chandola2007odreview}. Outliers often correspond to critical or novel cases due to data shift or other causes \cite{hong2024odreview}. Therefore, OD is crucial for reliable AI systems, helping them defer to human expertise when inputs lie outside the learned domain \cite{hong2024odreview}. Classical OD algorithms extract summary statistics that characterize the training distribution, against which a new input is compared \cite{kennedy2025-outlier, chandola2007odreview}. Recent approaches use deep learning and uncertainty measures. Particularly, vision–language models (VLM) offer rich joint embeddings that can provide semantic representations of images \cite{radford2021clip}. VLM such as CLIP has shown exceptional capability in downstream tasks, including outlier detection \cite{li2025odclip}. 

In this work, we showcase a hybrid anomaly detector in conjunction with two GenAI systems by BIOEMTECH.

\section{System Overview}

The proposed outlier detection pipeline is implemented in Obz AI. Features are modeled and logged by the Python library\footnote{\url{https://pypi.org/project/obzai}}. Integrated with our app\footnote{\url{https://obz.ai}}, a user can visualize features, detect outliers, and manage the project on the web dashboard.

\subsection{Modeling First Order Features}

To identify out-of-distribution (OOD) samples in preclinical imaging data, we investigated a suite of first-order statistical features (FOFs) from each sample. FOFs such as entropy, median, variance, and uniformity capture essential aspects of the data distribution while maintaining computational efficiency and interpretability.

To model the typical variation in the data, we fit a Gaussian Mixture Model (GMM) using the extracted FOFs from the training dataset \cite{pbml}. The GMM effectively learns the underlying distribution of in-distribution samples in the FOF space. Then, new samples in production are projected, and their likelihood under the GMM is evaluated. Samples whose likelihood falls below $\tau$ percentile of the GMM distribution are declared outliers. This threshold is empirically chosen to adjust for sensitivity and specificity.

\subsection{Visual-Language Embedding}

We use the CLIP (Contrastive Language-Image Pre-training) model to convert each image in the training dataset into a high-dimensional embedding vector \cite{radford2021clip, li2025odclip}. These embeddings serve as a compact yet information-rich representation of image content. Once the embeddings are obtained, we perform Principal Component Analysis (PCA) to identify a lower-dimensional subspace that captures the majority of variance within the training data \cite{jolliffe1986pca}.

The reconstruction loss for each sample is defined as the squared Euclidean norm between the original embedding vector $\mathbf{x}_i$ and its projection $\hat{\mathbf{x}}_i$ onto the subspace spanned by the top $r$ principal components: $\text{Loss}_i = \|\mathbf{x}_i - \hat{\mathbf{x}}_i\|^2$. A low reconstruction loss indicates that the sample is well explained by the distribution of the training data, as captured by $r$ PCs. To set the dimensionality $r$ for optimal outlier identification, we conduct a parameter sweep such that the proportion of outlier samples detected in the test dataset matches most closely the proportion found in the training dataset.

\section{Applications}
\subsection{Generate X-Ray Images from Photos}


\begin{figure}[t]
  \centering
    \includegraphics[width=.5\textwidth]{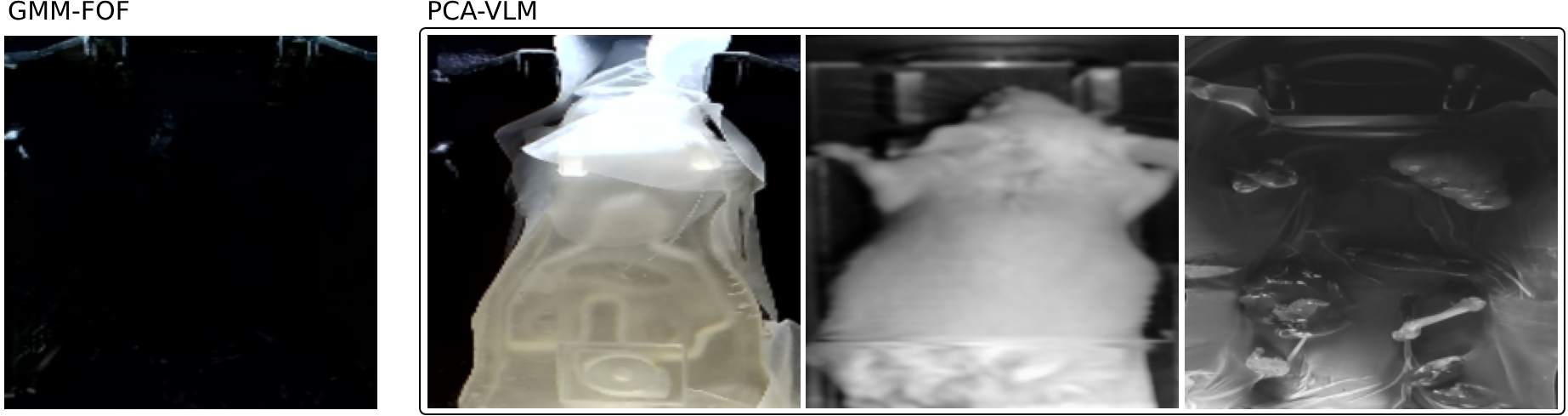}
    \caption{Example outliers from the Pose2Xray application}
    \label{fig:pose2xray}
    \vspace{-1em}
\end{figure}

This preclinical system introduces a two-stage GenAI for synthesizing morphological X-ray images of mice from standard photographic images. The pipeline first employs a Keypoint R-CNN model \cite{he2017mask}, trained on 1,294 annotated mouse images, to detect 18 anatomical keypoints per mouse and generate a corresponding pose image. Subsequently, we trained a Pose2Xray model on 1,428 pose/X-ray pairs to generate synthetic X-rays \cite{vasileios2025}. This GenAI function is incorporated into the \emph{eyes}\textsuperscript{\texttrademark} imaging systems by BIOEMTECH. Accurate representation of mouse anatomy is essential for the alignment and superimposition of these synthetic anatomical references onto the functional biodistribution data.


In an early model version, significant limitations led to frequent misalignments and inaccurate synthetic X-rays, especially under varying mouse models, imaging conditions, or non-standard sample types (e.g., phantoms). Incorporating the proposed OD allows for automated flagging and discarding of inaccurate predictions (\autoref{fig:pose2xray}), thereby improving trust in synthetic data and enabling wider adoption of AI-powered imaging tools across diverse laboratory settings. The integration of OD enabled the industrial viability of these GenAI systems, particularly valuable for settings with limited technical oversight or high-throughput imaging demands. With synthetic anatomy available immediately during acquisition, imaging time is significantly reduced, a critical factor given that small animals must remain under anesthesia, which should be minimized due to physiological and functional constraints.

\subsection{Generate Dose Maps from SPECT/CT Scans}


\begin{figure}[t]
  \centering
    \includegraphics[width=.5\textwidth]{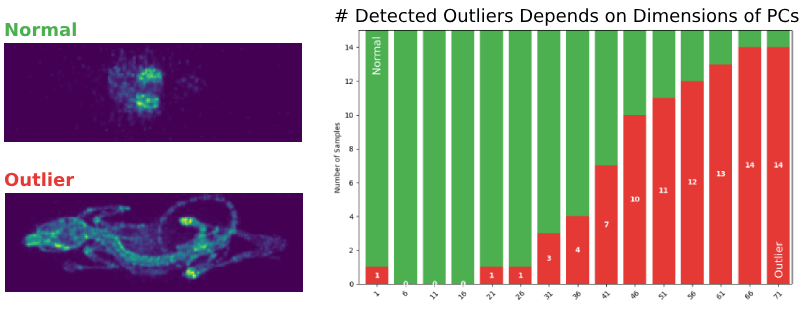}
    \caption{Outlier detection in DosimetrEYE}
    \label{fig:vox2vox}
    \vspace{-1em}
\end{figure}

We developed a Vox2Vox model \cite{vox2vox} to estimate 3D absorbed dose distributions from 2D SPECT/CT scans, which we call \emph{DosimetrEYE}. Conventionally, accurate dosimetry requires time-intensive and resource-demanding Monte Carlo (MC) simulations using 3D SPECT/CT data, limiting their practical use in preclinical imaging. By pairing planar images with simulated 3D dose maps as ground truth, this approach enables non-invasive, real-time dosimetry in small-animal studies, facilitating efficient organ-level dose assessment.


However, the DosimetrEYE model is trained on a small set of training data which are labor- and capital-intensive to obtain. Therefore, it is imperative to reject out-of-distribution samples before analysis to reduce inaccurate predictions. The proposed integration with outlier detection enables real-time quality control, helping establish a robust, integrated workflow where dosimetry is performed in parallel with imaging. This innovation not only streamlines operational efficiency and reduces computational overhead but also contributes to reduction of over 80\% animal sacrifice. This shift represents a transformative step toward ethical, scalable, and data-rich preclinical studies in the radiopharmaceutical industry.

\section{Conclusion}

In industrial applications of GenAI for preclinical imaging and dosimetry, outlier detection is essential for ensuring product reliability, scalability, and regulatory readiness. Integrating outlier detection directly into industrial GenAI pipelines to automatically identify anomalous or low-quality imaging data that could compromise model performance or lead to inaccurate dose predictions. This not only improves the robustness of synthetic data but also reduces the need for manual data inspection, accelerating validation cycles in high-throughput environments. By flagging edge cases early, outlier detection enables proactive quality control, supports data traceability, and builds trust in AI-assisted decision-making — key requirements for industrial adoption. Ultimately, this functionality transforms GenAI from a research tool into a dependable feature embedded in imaging products, supporting efficient development workflows and enhancing the reproducibility and regulatory acceptance of preclinical studies.

\begin{acks}
This work was funded by the SONATA BIS [2023/50/E/ST6/00694] from Narodowe Centrum Nauki; by the FFplus Innovation Study [1204; DOSIMETREYE], which is funded by the European High-Performance Computing Joint Undertaking (JU) under grant agreement No 101163317. The JU receives support from the European Union's Horizon Europe Programme.
\end{acks}

\bibliographystyle{ACM-Reference-Format}
\bibliography{citations}



\end{document}